\documentclass[lettersize,journal]{IEEEtran}
\usepackage{amsmath,amsfonts}
\usepackage{algorithmic}
\usepackage{algorithm}
\usepackage{array}
\usepackage[caption=false,font=normalsize,labelfont=sf,textfont=sf]{subfig}
\usepackage{textcomp}
\usepackage{stfloats}
\usepackage{url}
\usepackage{verbatim}
\usepackage{graphicx}
\usepackage{cite}
\usepackage{hyperref}
\usepackage{fancyhdr} 

\usepackage{fancyhdr}

\begin{document}

\title{Understanding Audiovisual Deepfake Detection:\\ Techniques, Challenges, Human Factors\\ and Perceptual Insights}

\author{IEEE Publication Technology,~\IEEEmembership{Staff,~IEEE,}
\thanks{This paper was produced by the IEEE Publication Technology Group. They are in Piscataway, NJ.}
\thanks{Manuscript received April 19, 2021; revised August 16, 2021.}}

\author{Ammarah Hashmi, Sahibzada Adil Shahzad, Chia-Wen Lin,~\IEEEmembership{Fellow,~IEEE}, Yu Tsao,~\IEEEmembership{Senior Member,~IEEE}, and Hsin-Min Wang,~\IEEEmembership{Senior Member,~IEEE}

\thanks{Ammarah Hashmi is with the Social Networks and Human-Centered Computing Program, Taiwan International Graduate Program, Institute of Information Science, Academia Sinica, Taipei 11529, Taiwan, and also with the Institute of Information Systems and Applications, National Tsing Hua University, Hsinchu 300044, Taiwan. (e-mail: hashmiammarah0@gmail.com).

Sahibzada Adil Shahzad is with the Social Networks and Human-Centered Computing Program, Taiwan International Graduate Program, Institute of Information Science, Academia Sinica, and also with the Department of Computer Science, National Chengchi University, Taipei 11529, Taiwan. (e-mail: adilshah275@iis.sinica.edu.tw).

Chia-Wen Lin is with the Department of Electrical Engineering and the Institute of Communications Engineering, National Tsing Hua University, Hsinchu 300044, Taiwan. (e-mail: cwlin@ee.nthu.edu.tw).

Yu Tsao is with the Research Center for Information Technology Innovation, Academia Sinica, Taipei 11529, Taiwan. (e-mail:
yu.tsao@citi.sinica.edu.tw).

Hsin-Min Wang is with the Institute of Information Science, Academia Sinica, Taipei 11529, Taiwan. (e-mail: whm@iis.sinica.edu.tw).

}}



\maketitle
\thispagestyle{fancy} 

\begin{abstract}
Deep Learning has been successfully applied in diverse fields, and its impact on deepfake detection is no exception. Deepfakes are fake yet realistic synthetic content that can be
used deceitfully for political impersonation, phishing, slandering, or spreading misinformation. Despite extensive research on unimodal deepfake detection, identifying complex deepfakes through joint analysis of audio and visual streams remains relatively
unexplored. To fill this gap, this survey first provides an overview of audiovisual deepfake generation techniques, applications, and their consequences, and then provides a comprehensive review of state-of-the-art methods that combine audio and visual modalities to enhance detection accuracy, summarizing and critically analyzing their strengths and limitations. Furthermore, we discuss existing open source datasets for a deeper understanding, which can contribute to the research community and provide necessary information to beginners who want to analyze deep learning-based audiovisual methods for video forensics. By bridging the gap between unimodal and multimodal approaches, this paper aims to improve the effectiveness of deepfake detection strategies and guide future research in cybersecurity and media integrity.
\end{abstract}


\section{Introduction}
\IEEEPARstart{T}{he} proliferation of smart digital devices such as mobile phones, laptops, tablets, and other digital gadgets, coupled with the accessibility of social media platforms, has promoted the exponential growth of multimedia content (images, videos, and audio) on the internet. This growth is further fueled by technological advances \cite{R152}, including various deep generative networks \cite{R3} \cite{R4}. However, this accessibility heightens the need for caution because it can lead to the prevalence of disinformation. Despite this, many people still stick to the trend of the antiquated phrase ``seeing is believing'' and share multimedia content without considering its authenticity or verifying its digital integrity. Deepfake technology, or sophisticated Artificial Intelligence (AI) models, enable deep learning (DL) tools to manipulate media (images, videos, and audio) to generate hyper-realistic fake content that deceives viewers. Deepfake is AI-generated media that has been deceptively altered by superimposing a source face in a video onto a target face, manipulating the speech in an audio clip, or both. The vast amount of data available online in the form of images, videos, and audio to train such models makes detecting such forgeries increasingly challenging.
The impact of deepfakes is critical because we still trust photographic and audio recording evidence. The emergence of realistic and subtle production tools makes fake content incredibly believable and harder to distinguish from genuine content \cite{R151}. The rapid spread of harmful and uncontrolled content from fake media has serious imminent impacts and reduces trust in journalism and news providers \cite{R12} \cite{R89}. Deepfake media content can be exploited to fuel political or religious tensions between countries \cite{R90}, spread misleading information or rumors between political parties \cite{R12} \cite{R91}, deceive the public \cite{R12}, engaging in revenge porn \cite{R91}, defame celebrities \cite{R91}, promote fraud and identity theft \cite{R13}, and create political chaos or publicity in a campaign \cite{R92}.

Generative Adversarial Networks (GAN) \cite{R3} and Variational Autoencoders (VAE) \cite{R4} are sophisticated DL models for generating counterfeit content. In GAN, the generator network and the discriminator network are the two main components, and these two networks are opposed to each other. The generator aims to generate plausible data, while the discriminator determines the real data from the fake data generated by the generator. Similarly, VAE is an unsupervised learning method consisting of encoder and decoder architectures. VAE is used to create high-quality, hyper-realistic fake content by merging and/or superimposing existing media (images or videos) onto source media for the purpose of deception.  

Currently, AI-synthesized videos are mainly divided into three different generation types \cite{R2} \cite{R8}. (1) Head puppetry/puppet master is a counterfeit video generation technique based on the target person animating like a puppet. (2) Face swap aims to generate a video of the target person by swapping the target person's face with that of the source person while retaining the same facial expression as the target person. (3) Lip-sync is another deepfake video generation method whose main goal is to transform a person's lips to be synchronized or consistent with the target audio. This technique tends to manipulate the lip region in such a way that the target of the attack appears to be saying things they never said in reality. 

In the past few years, immense progress in the field of automatic video editing and a great interest in face manipulation techniques have been noticed. Advances in manipulation tools and open-source codes allow even naive users to use deepfake technology like an expert in a few simple steps. This technological advancement has a wide range of positive applications in the fields of visual effects, photography, education, film industry, virtual reality, video games, cinema, and entertainment. However, it also poses significant challenges in terms of authenticity verification and prevention of malicious use.
To overcome these challenges, researchers have made many attempts and proposed DL-based unimodal forgery detection methods \cite{R14, R15, R16, R17, R20}. 

\begin{figure}[!t]
\centering
\includegraphics[width=3.5in]{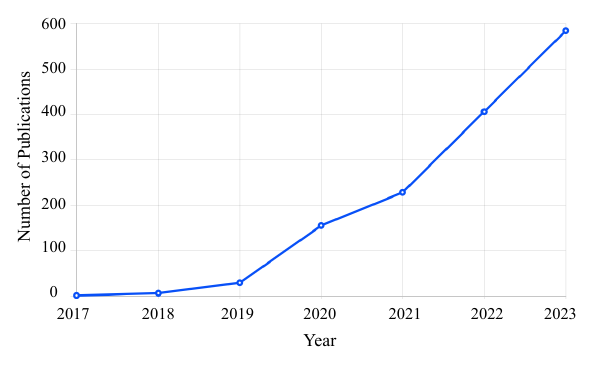}
\vspace{-20pt}
\caption{Volume of research into audiovisual deepfakes between 2017 and 2023.}
\label{Fig1}
\vspace{-8mm}
\end{figure}

The detection of visual manipulation in videos has long been the focus of researchers, while the identification of audio forgeries has often been overlooked. Recently, however, the trend of sound manipulation has grown rapidly along with visual alterations, leading to bimodal fabrication that enhances the authenticity of fake content and makes detection difficult \cite{R18, R19, R22, R23}. Fig. \ref{Fig1} highlights the research community's growing interest and concern in audiovisual deepfakes. The number of publications on audiovisual deepfakes has increased significantly in recent years, demonstrating both beneficial progress and growing concerns. The integration of multimodality is proven to be beneficial in various research fields \cite{R153, R154, R155}. Consequently, researchers have used various DL techniques that exploit audio and visual features for video forgery detection. Nonetheless, existing media forensics research is lacking in investigations that analyze methods for generating and detecting video deepfakes using audio and visual modalities. Table \ref{table1} lists an overview of relevant studies. Our study was strongly motivated by the lack of attention paid to audiovisual deepfakes in surveys, highlighting the urgent need to focus research on audiovisual deepfakes, including their generation, how to mitigate their harmful effects, and a summary of existing audiovisual deepfake detection methods.

\begin{table}[!t]
\centering
\small
\caption{Comparison of survey studies related to deepfake detection.}
\scalebox{0.9}{
\begin{tabular}{|m{1.6cm}|m{0.5cm}|m{6.2cm}|}
\hline
\textbf{Reference} &  \textbf{Year} & \textbf{Contribution} \\ \hline\hline

Verdoliva \cite{R31} & 2020 & A discussion of video deepfakes from a forensic perspective, with an emphasis on the limitations of current forensic detection methods. \\ \hline
Mirsky et al. \cite{R37} & 2021 & An in-depth analysis of field-specific generation techniques and a brief discussion of detection methods. \\ \hline
Yu et al. \cite{R32} & 2021 & A detailed analysis of forged video synthesis and detection techniques, with a focus on face manipulation. \\ \hline
Rana et al. \cite{R33} & 2022 & A comprehensive review of deepfake detection methods proposed during 2018-2020.  \\ \hline
Nguyen et al. \cite{R34} & 2022 & A comprehensive overview of deepfake generation and detection techniques and a discussion of challenges and future research directions in the field. \\ \hline
Masood et al. \cite{R35} & 2023 & An analysis of the generation and detection of audio and visual deepfakes and a discussion of datasets. \\ \hline
Mubarak et al. \cite{R36} & 2023 & An analysis of audio, visual, and text-based deepfakes, with a focus on detection methods.\\ \hline
\end{tabular}}
\label{table1}
\vspace{-6mm}
\end{table}

\begin{figure}[!b]
\vspace{-6mm}
\centering
\includegraphics[width=3.5in]{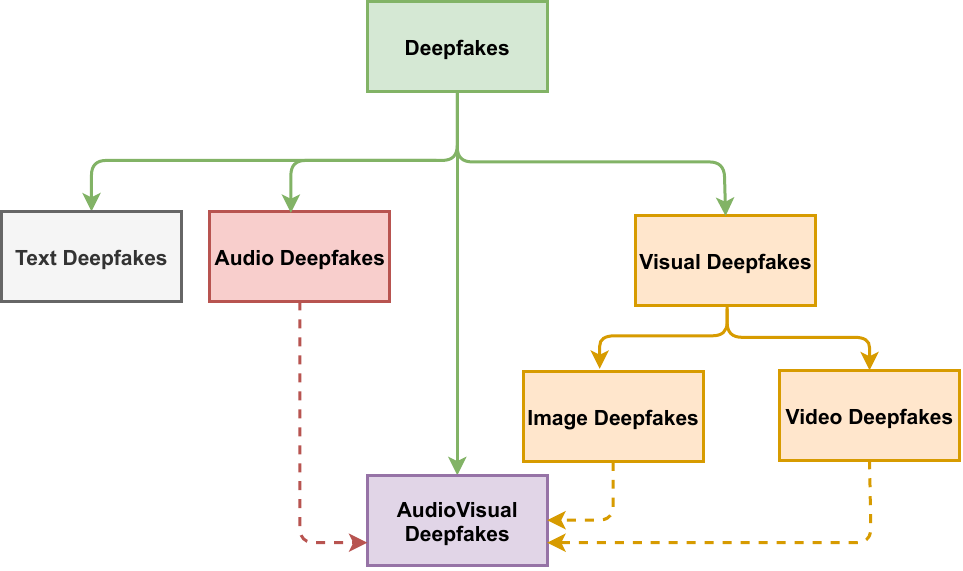}
\caption{Taxonomy of deepfakes.}
\label{Fig2}
\vspace{-6mm}
\end{figure}

Generally speaking, as shown in Fig. \ref{Fig2}, there are four types of deepfakes, namely text deepfakes, audio deepfakes, visual deepfakes, and audiovisual deepfakes. Audiovisual deepfakes are a combination of acoustic and visual manipulation that can enhance manipulated videos to make them look more believable, and have received a lot of attention in recent years. This study specifically provides an in-depth review of the latest audiovisual deep learning solutions to improve the detection of challenging video deepfakes. 
To the best of our knowledge, we are the first to perform a comprehensive analysis of existing DL-based methods that exploit audio and visual manipulations in videos for automatic deepfake detection. Our important contribution also includes a comprehensive discussion of publicly available datasets relevant to this task. The main contributions of our work are as follows:

\begin{itemize}
\item We provide an unprecedented survey that systematically analyzes key detection and generation methods for audiovisual deepfakes, with a special emphasis on automatic video deepfake detection methods.
\item We highlight the challenges, limitations, and human perception in the field of audiovisual deepfake detection. Furthermore, we outline research directions for future developments in this field.
\item We summarize and present publicly available datasets that can be used to train multimodal/audiovisual deepfake detectors.
\end{itemize}

The remainder of this paper is organized as follows. Section~\ref{sec:Deepfake Categories} introduces different types of deepfakes. Section~\ref{sec:Approaches to Video Deepfake Detection} discusses video deepfake detection methods. Section~\ref{sec:Classification of Audiovisual Deepfake Detection Methods} classifies detection methods that exploit visual and acoustic streams. In Section~\ref{sec:Datasets for Audiovisual Deepfake Detection}, we review publicly available datasets for audiovisual deepfake detection. Section~\ref{sec:Performance Metrics and Evaluation} presents performance metrics and evaluation.
Section~\ref{sec:Human Perception and Psychological Impact in Audiovisual Deepfakes} examines human perception of audiovisual deepfakes. Section~\ref{sec:Current Challenges and Future Directions} discusses several aspects of the deepfake challenge and potential research directions. 
Finally, Section~\ref{sec:Conclusion} concludes this survey.

\section{Deepfake Categories}
\label{sec:Deepfake Categories}
Deepfake is a type of AI-generated fake hyper-realistic media content that involves the manipulation in acoustic and/or visual modalities, making it difficult to distinguish between real and fake. This study focuses on audiovisual deepfakes, which combine audio and visual manipulations. In this section, we provide a brief overview of audiovisual deepfake generation methods and their components.

\subsection{Audio Deepfakes}
Audio deepfakes refer to voices that have been digitally altered but sound real.
Convincing audio deepfakes are usually achieved through AI techniques or DL models, such as GANs or VAEs. MelGAN~\cite{R24} and WaveGAN~\cite{R25} are well-known examples. The former is a generative model for raw audio, and the latter is a generative adversarial network for conditional waveform synthesis. Contemporary breakthroughs in audio deepfakes have increased the threat to voice interfaces, contributing to criminal activity and cybercrime, thus raising concerns about their misuse. Audio deepfake methods fall into three categories, as described below. 

\begin{figure}[!b]
\vspace{-6mm}
\centering
\includegraphics[width=3.0in]{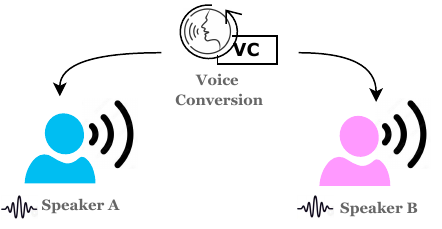}
\caption{Diagram of voice conversion.}
\label{Fig3}
\vspace{-2mm}
\end{figure}

\subsubsection{Voice Conversion}
Voice Conversion (VC) involves converting the speech signal of a first speaker (called the source) to match the voice of a second speaker (called the target), while preserving the original linguistic content. Natural speech is the input to the VC system, whose purpose is to change the timbre and prosody of the source to that of the target. Fig. \ref{Fig3} shows a schematic diagram of voice conversion. Audio deepfakes can be used as independent clips to high impact, or they can be combined with visual deepfakes to produce audiovisual fabricated content. MelGAN-VC \cite{R29}, an extension of MelGAN \cite{R24}, is a voice conversion and audio style transfer method applied to arbitrarily long samples. This method uses mel-spectrograms for voice conversion or style transfer through MelGAN. StarGAN \cite{R30} is a non-parallel many-to-many VC method based on star generative adversarial networks that can convert any source voice into the voice of any target speaker. VoiceLoop \cite{R26} is a DL-based VC method that converts speech from source to target through a phonological loop. SINGAN  \cite{R27} is a GAN-based singing voice conversion model that converts the singing voice of the source singer into the singing voice of the target singer.

\subsubsection{Text-to-Speech (TTS)}
\begin{figure}[!h]
\vspace{-4mm}
\centering
\includegraphics[width=3.0in]{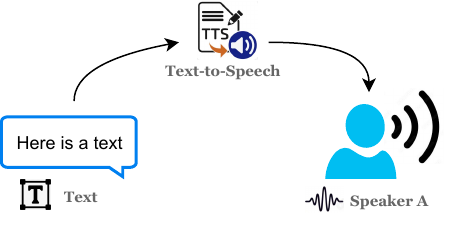}
\caption{Diagram of text-to-speech.}
\label{Fig4}
\vspace{-2mm}
\end{figure}

 Text-to-Speech (TTS), also known as Speech Synthesis (SS), is the artificial generation of human speech using software or hardware systems. Fig. \ref{Fig4} shows a schematic diagram of TTS. TTS models typically follow two steps to convert written text into human-like speech: text processing and speech generation. This technology can be used to create fake audio messages or impersonate voices for threatening purposes, but can also be useful for text-reading or personal assistants that provide different accents and voices than the pre-recorded human voice. Deep Voice \cite{R28} is a neural network-based synthesis model that generates text-to-speech in real time. Traditional TTS systems struggle to replicate natural flow and cannot mimic human speech. Recently, the quality of synthesized audio has been significantly improved through end-to-end models, such as Variational Inference with Adversarial Learning for End-to-End Text-to-Speech (VITS) \cite{R38} and FastDiff-TTS \cite{R39}. Other DL-based methods include WaveNet \cite{R40}, Deepvoice \cite{R41}, Tacotron \cite{R42}, and NaturalSpeech \cite{R43}.

\begin{figure}[!b]
\vspace{-6mm}
\centering
\includegraphics[width=3.0in]{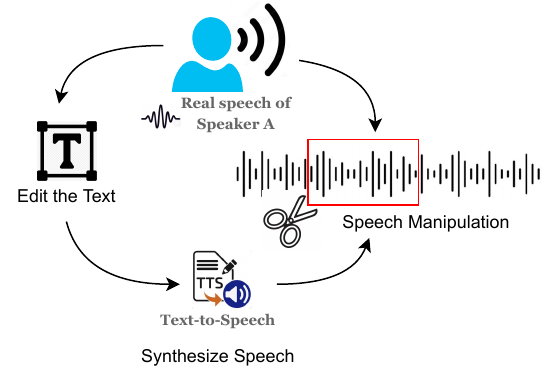}
\caption{Schematic diagram of the generation of partial audio deepfakes.}
\label{Fig5}
\end{figure}

\subsubsection{Partial Audio Deepfakes}
In contrast to full audio deepfakes, partial audio deepfakes \cite{R82} involve using AI techniques to alter specific parts of the original clip while maintaining the overall authenticity of the recording, rather than synthetically generating the entire audio. This complex selective alteration of specific segments of an audio recording makes detecting tampering more challenging than fully synthetic audio deepfakes. Fig. \ref{Fig5} shows the generation diagram of partial audio deepfakes. Partial deepfakes of audio have a range of positive applications in entertainment and media, education, and accessibility, such as correcting errors in recordings; however, the technology's negative applications cannot be ignored, such as misinformation and disinformation, fraud and impersonation, and manipulation and defamation.

\subsection{Visual Deepfakes}
Visual deepfakes refer to images or videos that have been digitally altered by DL technology but look realistic. This technology can alter facial expressions, gestures, lip movements, and body movements, or seamlessly superimpose the source person's face onto the target person's body, and easily deceive the viewer. Visual deepfakes have widespread applications in fields, such as education, gaming, and entertainment, but they pose considerable risks by leading to misinformation, defamation, and privacy violations. Five types of visual manipulation fall into this category, each of which is discussed separately in this section.

\subsubsection{Face Swap}
\begin{figure}[!h]
\vspace{-2mm}
\centering
\includegraphics[width=3.0in]{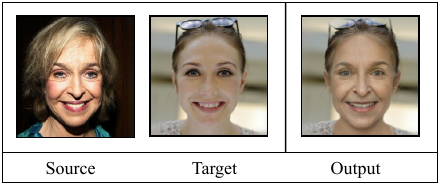}
\caption{Face swap example \cite{R129}.}
\label{Fig6}
\vspace{-2mm}
\end{figure}

Face swap \cite{R44}, a prominent visual manipulation technique, utilizes a generative model to seamlessly replace the target face with the source's identity in images or videos. Fig. \ref{Fig6} presents an example generated using the face swap technique. In certain scenarios, traditional methods \cite{R45,R46,R47} may fail to capture the expression depicted in the original face image, or sometimes result in an unnatural appearance. However, recent advances in DL have led to automated methods \cite{R48} \cite{R49} capable of producing plausible multimedia content. This technology has been integrated into various applications, such as FaceSwap \cite{R49}, DFaker \cite{R50}, DeepFaceLab \cite{R51}, Deepake-tf \cite{R52} and FaceSwapGAN \cite{R53}, which use deep neural networks to achieve realistic face-swapping results.

\subsubsection{Face Generation}
\begin{figure}[!t]
\vspace{-2mm}
\centering
\includegraphics[width=3.0in]{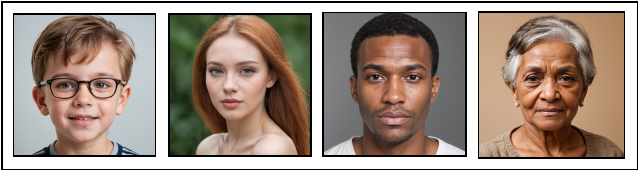}
\caption{Examples of face generation.}
\label{Fig7}
\vspace{-6mm}
\end{figure}

Face generation, also known as face synthesis \cite{R54}, creates realistic human faces that are completely fictional and do not correspond to any real identities using generative models such as GANs \cite{R3} and VAEs \cite{R4}. Fig. \ref{Fig7} shows some examples generated using this technique. This technology offers valuable applications, such as automatic character creation in the video gaming and 3D face modeling industries. However, it can also be used for malicious purposes, such as impersonating or spreading disinformation on social media \cite{R55}. Starting with the initial low-resolution imagery \cite{R3}, the quality has improved dramatically over time. Advanced models such as StyleGAN \cite{R56}, ProGAN \cite{R57}, StyleGAN2 \cite{R58}, TP-GAN \cite{R59}, SAGAN \cite{R60}, and BigGAN \cite{R61} can now produce high-resolution content.

\subsubsection{Reenactment}
\begin{figure}[!h]
\vspace{-2mm}
\centering
\includegraphics[width=3.0in]{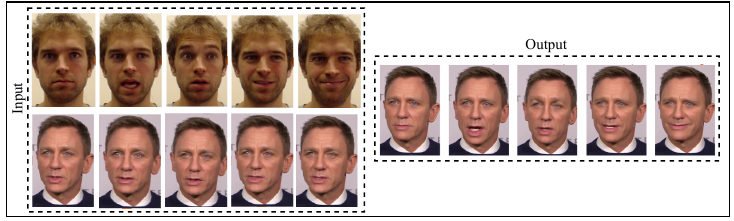}
\caption{Reenactment example.}
\label{Fig8}
\vspace{-2mm}
\end{figure}

Face reenactment, also known as puppet-master, is a technique for transferring facial expressions or body movements from a source video to a target video, as shown in Fig. \ref{Fig8}. Unlike face swap, this technique is designed to control and manipulate the target's expressions and movements to match those of the source, including gaze direction. Face reenactment has many applications in entertainment, virtual reality, and telepresence, but its potential for misuse can be harmful. Activists may creat deceptive videos that manipulate a person's expressions and movements to falsely portray them as engaging in behavior or speech that they never did. Some publicly available tools that utilize tracking and reenactment techniques to demonstrate facial expression and motion transfer include Face2Face \cite{R10}, FSGANv2 \cite{R62}, MarioNETte \cite{R63}, DeepFaceLab \cite{R51}, Pix2pixHD \cite{R64}, and ReenactGAN \cite{R65}.

\begin{figure}[!b]
\vspace{-6mm}
\centering
\includegraphics[width=3.0in]{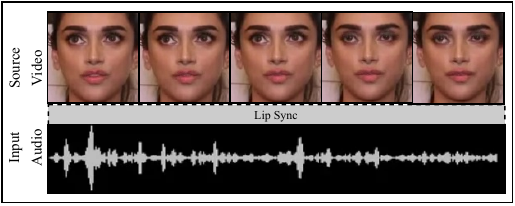}
\caption{Lip sync example.}
\label{Fig9}
\vspace{-4mm}
\end{figure}

\subsubsection{Lip Sync}
Lip sync involves changing a person's lip movements to match pre-recorded audio clips, ensuring accurate and convincing speech alignment. Fig. \ref{Fig9} shows an example of lip sync. This is a complex process, as the appearance and movement of the lower face, lip region, and surrounding areas are crucial to achieving the goal, while producing precise lip movements and expressions is also important to effectively convey the message. Lip sync has a variety of positive uses, including applications in forensic analysis, speech recognition research, and film production, where it ensures natural synchronization of dialogue with the audio soundtrack. However, it can have negative consequences when used to generate deceptive content for malicious purposes, such as spreading false information, defamation, and manipulating public opinion. The authors in \cite{R66, R68, R69} proposed techniques for lip synchronization. Other popular examples include Speech2Vid \cite{R67}, Wav2lip \cite{R70}, LipGAN \cite{R71}, and Vdub \cite{R72}.

\subsubsection{Facial Manipulation}
\begin{figure}[!h]
\vspace{-2mm}
\centering
\includegraphics[width=3.0in]{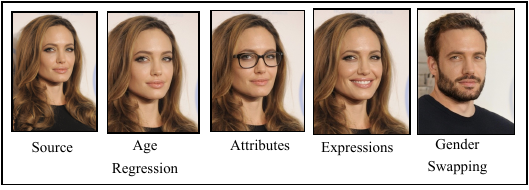}
\caption{Examples of facial manipulation.}
\label{Fig10}
\vspace{-6mm}
\end{figure}

Facial manipulation aims to alter or synthesize facial features in images and videos to create realistic but counterfeit representations. This technology leverages advanced algorithms and DL techniques \cite{R14} to modify expressions, age, gender, and even the entire facial structure to create highly realistic or completely fictional portraits, as shown in Fig. \ref{Fig10}. It has positive applications in various fields including entertainment, media, education and training, but its negative potential raises significant ethical and security issues. FaceApp \cite{R73}, SC-FEGAN \cite{R74}, AttGAN \cite{R75}, and Openart \cite{R76} are practical examples of manipulation tools showcasing advanced capabilities for altering and generating realistic facial images for entertainment purposes. 

\subsection{Text Deepfakes}
Text deepfakes are text that is artificially generated or altered using advanced DL or AI techniques that produce highly realistic and contextually appropriate text. These sophisticated algorithms represent powerful and versatile applications of AI, capable of generating articles, social media posts, conversations, and other forms of written communication that are often indistinguishable from content written by humans. However, they can be misused, so a balanced approach that combines robust detection methods, ethical guidelines, and public awareness is needed. These deepfakes fall into two categories, each with unique characteristics. Both types are reviewed separately in this section.

\subsubsection{Synthetic Text}
Synthetic text is generated by machines using advanced AI and DL techniques to imitate human writing with great accuracy. This type of text deepfakes are created using sophisticated natural language generation (NLG) models, such as GPT-3, GPT-4, and their successors \cite{R77} \cite{R78}. Large Language Models (LLM) are trained on large amounts of human writing and generate coherent and contextual text based on a given prompt, making it hard to discern between human and synthetic text \cite{R78}. This technology has a variety of positive applications \cite{R81}, including automated content generation for customer service, educational tools, and creative industries. However, the potential for misuse, such as spreading misinformation \cite{R79}, fake news articles, and conducting fraud, raises significant ethical and security concerns \cite{R80}. 

\subsubsection{AI-Powered Bot-Generated Text}
AI-powered bot-generated text refers to content created by automated systems or bots that use AI models to interact online. They excel at generating responses, posts, and other forms of communication that effectively mimic human behavior. AI empowers these bots to engage in social media discussions, comment on issues, and even generate news articles, often without being noticed by casual observers. The integration of AI in bots provides a variety of beneficial applications that boost the efficiency and engagement on digital platforms. However, as technology evolves, the potential misuse of this deceptive approach poses significant challenges to the responsible use of AI in online interactions. 

\subsection{Audiovisual Deepfakes}
\begin{figure}[!h]
\vspace{-4mm}
\centering
\includegraphics[width=3.0in]{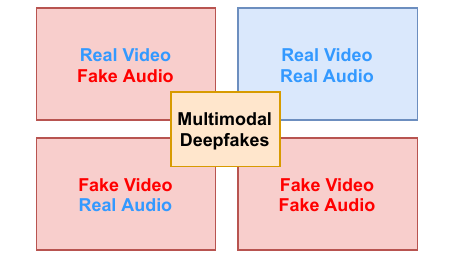}
\vspace{-10pt}
\caption{Four video categories based on acoustic and visual manipulations within them.}
\label{Fig11}
\vspace{-2mm}
\end{figure}

Audiovisual deepfakes refer to convincing and compelling fabricated videos created by manipulating the acoustic and visual streams of a video. Bimodal manipulation produces sophisticated deepfake videos that seamlessly blend fabricated video and audio content beyond single-modal manipulation. The misuse of audiovisual deepfakes poses significant risks, as they have the potential to spread misinformation, defamation, and other malicious activities. We discussed single-modal manipulation, namely audio and video deepfakes, in the previous two sections. This section focuses on audiovisual deepfakes that combine the two. Videos are divided into four categories based on acoustic and visual manipulations within them, as shown in Fig. \ref{Fig11}.

\subsubsection{Fake Video and Fake Audio (FVFA)}
\begin{figure}[!h]
\vspace{-2mm}
\centering
\includegraphics[width=3.0in]{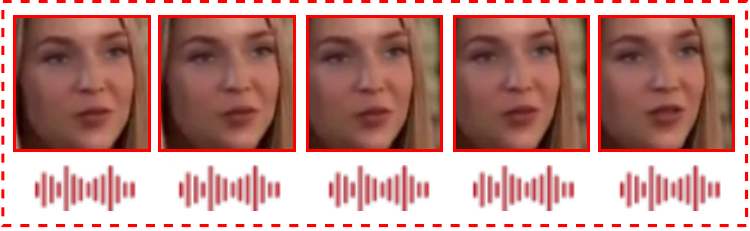}
\caption{Example of the Fake Video and Fake Audio (FVFA) category }
\label{Fig12}
\vspace{-4mm}
\end{figure}

Unlike partial deepfakes, the Fake Video and Fake Audio (FVFA) category \cite{R93} involves the artificial generation of both audio and visual components. It is the most comprehensive and advanced audiovisual deepfake, capable of fabricating complete acoustic and visual messages, often depicting fictional scenes that never happened. DL models, such as GANs, can generate high-quality synthetic visuals and produce realistic facial expressions, movements, and interactions that are coherent with the rest of the visual content. At the same time, synthetic audio that aligns with the video content is generated through TTS systems or voice cloning technology. These methods generate a highly persuasive audio track by analyzing the phonetic patterns of the target voice. Fig. \ref{Fig12} shows an example of FVFA.

\subsubsection{Real Video and Fake Audio (RVFA)}
\begin{figure}[!h]
\vspace{-2mm}
\centering
\includegraphics[width=3.0in]{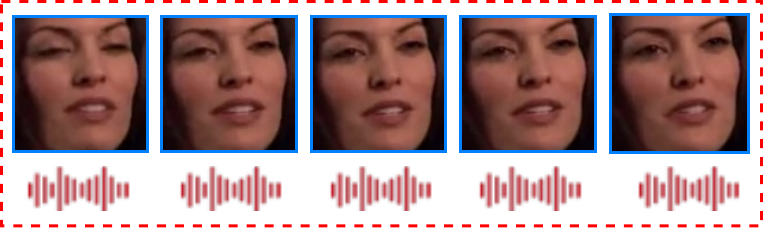}
\caption{Example of the Real Video and Fake Audio (RVFA) category.}
\label{Fig13}
\vspace{-4mm}
\end{figure}

The Real Video and Fake Audio (RVFA) category \cite{R93} involves combining authentic video with synthetically generated audio recordings. Fig. \ref{Fig13} shows an example of RVFA. AI and DL techniques alter the auditory message while preserving the visual content, making it appear as if someone in the video is saying something they never actually said. It is particularly effective in situations where altering the audio can significantly change the perceived message or context of an event (such as political manipulation or the creation of defamation fake news videos). The video is often based on real events or recordings and remains intact. The synthesized audio is created using a voice cloning or TTS system to ensure alignment with the lip movements and expressions of the person in the video.

\subsubsection{Fake Video and Real Audio (FVRA)}
\begin{figure}[!h]
\vspace{-2mm}
\centering
\includegraphics[width=3.0in]{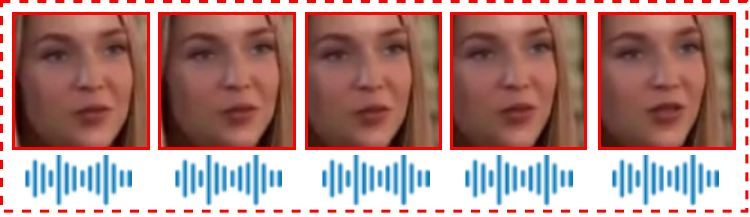}
\caption{Example of the Fake Video and Real Audio (FVRA) category.}
\label{Fig14}
\vspace{-4mm}
\end{figure}

The Fake Video and Real Audio (FVRA) category involves \cite{R93} the artificial manipulation of the visual track, while the sound track remains real and intact, as shown in Fig. \ref{Fig14}. This category is particularly effective in situations where altered visual scenes are combined with genuine audio to significantly change the perceived message or context of an event (such as the creation of fake news videos, defamation, or impersonation). State-of-the-art visual methods are used to produce realistic visual effects that are consistent and synchronized with real audio to present a coherent and realistic audiovisual impression.

\subsubsection{Real Video and Real Audio (RVRA)}
\begin{figure}[!h]
\vspace{-2mm}
\centering
\includegraphics[width=3.0in]{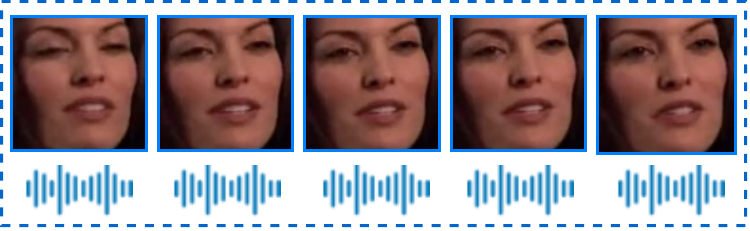}
\caption{Example of the Real Video and Real Audio (RVRA) category.}
\label{Fig15}
\vspace{-4mm}
\end{figure}

The acoustic and visual tracks in the Real Video and Real Audio (RVRA) category \cite{R93} remain unaltered and are derived from authentic video and audio recordings, preserving the originality and integrity of both modalities. Fig. \ref{Fig15} shows an example of RVRA. Technically, this category does not fall within the definition of deepfakes, but when used deceptively, it can be just as misleading as synthetic content.

\section{Video Deepfake Detection Methods}
\label{sec:Approaches to Video Deepfake Detection}
The field of video deepfakes is booming due to cutting-edge generative technology. At the same time, detection methods have emerged as crucial tools in combating these deepfakes. This section explores various techniques for detecting video deepfakes, divided into traditional methods based on visual cues and audio-based methods.

\subsection{Traditional Methods Based on Visual Cues}
Traditional methods for video deepfake detection heavily rely on visual irregularities or cues that arise while altering visual content \cite{R94}, such as inconsistencies, anomalies, and artifacts within video frames \cite{R95}. These methods can identify subtle anomalies and are broadly divided into two subcategories, both of which are discussed separately below.   

\subsubsection{Frame Analysis}
Frame analysis focuses on examining individual video frames to identify visual artifacts or inconsistencies such as unnatural lighting \cite{R96}, irregularities in shadows \cite{R97} \cite{R98}, and distortions around the edges of manipulated regions that indicate tampering \cite{R99}. Many techniques fall into this category, each focusing on different aspects of visual content. Pixel-level analysis \cite{R94}, error level analysis (ELA) \cite{R100}, compression artifacts \cite{R101}, edge detection \cite{R102}, and frequency domain analysis \cite{R102} are some examples of frame analysis techniques.

\subsubsection{Inconsistencies in Facial Movements}
Inconsistencies in facial movements are a major indicator of deepfake manipulation, which along with expressions are difficult to perfectly replicate. Conventional methods exploit these irregularities, such as unnatural or inconsistent facial movements \cite{R106}, blinking patterns \cite{R103}, eye movements \cite{R104}, expression transitions \cite{R105}, or synchronization issues, to spot deepfakes. Temporal consistency checks \cite{R107}, landmark-based methods \cite{R108}, blink rate analysis \cite{R109} and frame-by-frame analysis \cite{R110} are some examples of such techniques.

\subsection{Audio-Based Methods}
Audio-based deepfake detection methods emphasize analyzing audio aspects to reveal signs of tampering in videos. These methods are important traditional techniques for finding inconsistencies between audio and visual streams or detecting anomalies in the audio itself. Voice analysis and speech-content discrepancies are two main categories of audio-based methods for ensuring the authenticity of audiovisual content.

\subsubsection{Voice Analysis}
Voice analysis techniques detect anomalies by analyzing acoustic properties such as speech patterns, voice characteristics, and linguistic nuances to identify deviations from natural audio characteristics \cite{R111} \cite{R112}. Analyzing audio frequency, pitch, and temporal characteristics provides valuable clues for identifying fake multimedia content. Spectrogram analysis \cite{R113} and voice biometrics \cite{R114} are examples of voice analysis techniques.

\subsubsection{Speech-Content Discrepancies}
The alignment between the audio and visual components in deceptive content is often imperfect. When the movement of the lips does not align precisely with the words spoken, there can be a noticeable mismatch, indicating tampering. This technique uncovers inconsistencies that exhibit manipulation by analyzing synchronization between audio and visual components \cite{R20} \cite{R23}. Lip-sync detection \cite{R18} and Phoneme-Viseme Analysis \cite{R115} are well-known examples.

\section{Audiovisual Deepfake Detection Methods}
\label{sec:Classification of Audiovisual Deepfake Detection Methods}
Deepfake videos threaten personal privacy and social security. Various strategies have been proposed to overcome this risk, with early attempts mainly focusing on single modality, while current detection methods focus more on multimodal features to identify manipulation in videos. Multimodal approaches combine the strengths of multiple modalities such as audio and visual data to enhance the accuracy and reliability of deepfake detection systems. These methods can effectively discover more subtle manipulations by integrating information from both modalities that single-modal methods may ignore. This systematic literature review examines existing video deepfake detection methods that use both audio and visual streams to detect forgeries in videos. We classify them into different categories and provide detailed information and discussion of the methods in each category later in this section. 

\subsection{Synchronization-Based Methods}
Synchronization between audio and visual streams is a critical aspect of multimodal deepfake detection and is essential to creating a seamless and coherent viewing experience \cite{R83} \cite{R84}. These synchronization methods are designed to synchronize audio and visual streams in time, specifically lip movements, expressions, and overall timing. Since speech, lip movements and facial expressions are naturally aligned in authentic videos, any discrepancy between audio and video streams indicates that the content has been tampered with or degraded, making synchronization techniques a critical tool in the fight against audiovisual fraud. 

In \cite{R115}, phoneme-viseme mismatches are exploited. The authors claim that producing sounds like ``M'', ``B'', and ``P'' requires complete lip/mouth closure, while deepfake techniques often fail to imitate the dynamics of lip sequences, making it a clue for forgery detection.

In addition to modality-specific embeddings, the study in \cite{R23} also integrates facial and speech emotion cues using state-of-the-art emotion recognition models to enhance the effectiveness of audiovisual forgery detection.

The authors of \cite{R22} exploit the inherent synchronization between video and audio streams to effectively identify inconsistencies and ultimately help spot manipulation in videos to improve deepfake detection. The proposed approach is a two-plus-one joint detection framework that jointly models audio and video, with one stream processing video frames and the other processing audio waveforms. The final deepfake prediction is produced by combining the outputs of these two streams, exploiting the inherent synchronization between them to detect misalignments between the two modes to identify forgeries in videos. Experiments demonstrate the high generalization ability of the proposed approach and show that the proposed joint audiovisual approach outperforms unimodal methods. 
    
The authors of \cite{R87} exploit audiovisual dissonance known as Modality Dissonance Score (MDS) to expose the video forgery. They believe that altering either modality can lead to noticeable dissonances, such as loss of lip synchronization or unnatural facial movements, which can be used as indicators that reveal altered material. Furthermore, their approach provides insight into altered video portions by locating segments in the video that exhibit signs of manipulation.

The study in \cite{R85} addresses the limitations of previous work that focused primarily on facial feature analysis and ignored the audio component and the broader context of audiovisual synchronization. The authors combined a phoneme-based audiovisual matching strategy and proposed the Audio-Visual Coupling Model (AVCM), which aims to capture the complex relationship between mouth movements (visual) and corresponding speech segments (audio) by measuring the similarity between them.  Experiments show the high performance of their approach compared to cutting-edge techniques, highlighting its robustness to detecting deepfakes by exploiting the intrinsic synchronization of audio and video streams. 

A unified audiovisual learning framework called AVoiD-DF is proposed in \cite{R7}, which effectively captures cross-modal and intra-modal inconsistencies or discrepancies between audio and visual content by jointly utilizing audio and visual features. This joint learning approach highlights the importance of simultaneously using audio and visual cues to substantially reduce misclassification rates, especially in cases where one modality has been tampered with.

The authors of \cite{R18} present an interesting work that focuses on the discrepancies between audio and visual components, specifically lip sync mismatch. Specifically, their proposed method detects inconsistencies between the lip sequence extracted from the video and the synthetic lip sequence generated from the audio using the wav2lip model. Furthermore, a pretrained convolutional neural network (CNN)-based lip-reading model is used to compare the extracted lip sequence with the synthetic lip sequence to distinguish synchronous and asynchronous audiovisual pairs, which helps distinguish real and fake videos. This work was further extended in \cite{R131} by utilizing audiovisual features extracted directly from the transformer-based multimodal model AV-HuBERT to identify inconsistencies between audio and visual components, thereby eliminating the wav2lip generation model.

\subsection{Feature Fusion Methods}
Feature fusion techniques leverage the complementary strengths of various features to enhance the accuracy of prediction models by combining features extracted from multiple modalities to create comprehensive and robust media representations. This holistic approach can significantly improve the accuracy of deepfake detectors, allowing them to capture more complex and informative representations of the underlying patterns.

The study in \cite{R86} highlights the importance of feature fusion by using jointly learned representations of audio and visual modalities to identify forgeries in videos. The authors propose a specialized approach called Multimodaltrace, which leverages the rich interdependencies between audio and visual signals in video content. This framework efficiently evaluates the audio and visual modalities by using a combination of channel extractors and mixers. 

Another recent study in \cite{R135} proposes a dual transformer model called AVT2-DWF with a dynamically weighted fusion strategy that calibrates its focus on each modality based on the relative strength and consistency of cues, thereby improving its detection of subtle or complex multimodal forgeries.

The authors of \cite{R136} propose an innovative two-stage cross-modal learning method called Audio-Visual Feature Fusion (AVFF) to distinguish authentic video content from fabricated video content by focusing on audiovisual coherence. Their method identifies audio and visual perturbations by using distant audio and visual features and examining unimodal and cross-modal embeddings. 

In \cite{R137}, the authors introduce a novel one-class learning method by combining acoustic and visual features. This multi-stream fusion approach uses various fusion strategies (early, intermediate, and late) to effectively integrate audio and visual signals and surpasses single-modality detection models, demonstrating robust performance across diverse scenarios. 

The study in \cite{R138} utilizes audio and visual modalities by extracting features and passing them to the multimodal model that learns nuanced differences across both domains. This framework explored two fusion strategies, feature fusion, and score fusion, demonstrating that simultaneously exploiting acoustic and visual features is effective in intra-domain and cross-domain testing environments.

To address heterogeneous feature fusion limitation, a novel multi-modal attention framework based on recurrent neural networks (RNNs) is proposed in \cite{R148} to exploit contextual information for audio-visual deepfake detection. The proposed approach utilizes attention mechanisms on multi-modal, multi-sequence representations, identifying the most relevant features across modalities to improve deepfake detection and localization.

\subsection{Ensemble Methods}
Ensemble approaches have proven to be highly beneficial in improving the accuracy of detectors in identifying audiovisual forgeries in videos, as they are able to take advantage of multiple models. Compared with a single model, integrating different models can effectively account for variations in forgery technology. 

The study in \cite{R140} presents two ensemble methods for audiovisual deepfake detection. The first method is soft-voting, which combines model predictions by averaging their probabilities, while the second method is hard-voting, which applies majority voting based on model outputs.

The authors of \cite{R133} introduce AVFakeNet, an ensemble architecture that uses visual and audio features to distinguish authentic videos from fake ones. Their model leverages a Dense Swin Transformer to handle bimodal complexity, extract and fuse audio and visual features, and ensure that the cross-modal relationship between visual and audio signals is effectively captured.

Similarly, the study in \cite{R19} introduces an integrated network consisting of three CNN-based networks. The authors utilize ensemble learning to integrate multiple modality-centric models to detect audiovisual forgeries in videos. Dedicated audio-only, video-only, and audiovisual networks make separate predictions, and a voting mechanism then fuses these predictions to make the final prediction. In the extension work called AVTENet \cite{R88}, the CNN architectures in audio-only, video-only, and audiovisual networks respectively were replaced by dedicated pretrained Transformer models, thereby achieving better performance.


\subsection{Temporal Analysis-Based Methods}
To address the limitation of simple feature fusion techniques, the method proposed in \cite{R130} utilizes dual networks to extract temporal features from both modalities. Specifically, audio and video modules are proposed to predict acoustic and visual temporal features and match them with reference features to capture temporal inconsistencies in audio and video modalities. A contrastive objective function is employed to maximize the difference between authentic and spoofed modalities, significantly enhancing the discriminative ability of the approach to classify genuine and forged instances. 

For temporal forgery detection (TFD), a multi-dimensional contrastive loss is introduced in \cite{R132} to help forensic models exploit temporal inconsistencies by constraining extracted embeddings. Experiments on the LAV-DF \cite{R9} dataset show the effectiveness of the proposed method. 

In \cite{R134}, the authors proposed a hybrid approach that utilizes pretrained models to extract spatial, spectral, and temporal features to distinguish real and fake video content. 

A self-supervised learning (SSL) based approach is employed in \cite{R21} to capture a rich representation based on temporal synchronization between speech and facial movement across the frames. The representation learned by the SSL network is then fed into a temporal classifier network to judge the authenticity of video content.

\subsection{Other Methods}
Boundary Aware Temporal Forgery Detection (BA-TFD) \cite{R9}, is introduced to address temporal forgery localization task. BA-TFD is a 3D Convolutional Neural Network (3DCNN)-based model proposed to localize forgery through three guiding objective functions:  contrastive, boundary matching, and frame classification. Later on, BA-TFD+ \cite{R141} improves the baseline BA-TFD approach by replacing its backbone with a Multiscale Vision Transformer coupled with a refined training process by an additional multimodal boundary matching loss function. 

MIS-AVoiDD \cite{R143} jointly utilizes modality-invariant and modality-specific features by focusing on shared and unique features across the modalities.

The method in \cite{R149} introduces a multimodal contrastive learning (MCL) approach to detect forgeries by capturing both intra-modal and cross-modal forgery cues. It aligns representations from audio, frames, and video using a cross-modal contrastive strategy and distills frame knowledge to the video network to strengthen forgery detection without extra computational cost. Additionally, a noise-based feature augmentation (NFA) module further enhances generalization by adaptively perturbing audio-visual features. 

DF-TransFusion \cite{R150} employs cross-attention between lip movements and audio signals to detect lip-sync inconsistencies, while a self-attention mechanism targets facial features to reveal subtle visual manipulations. This integrated approach enhances detection accuracy by identifying cross-modal discrepancies, enabling more effective detection of complex deepfake manipulations.

For real-world applications, \cite{R146} proposes a modality agnostic-based method to address the missing modalities issue in the multimodal deepfake detection task. This method facilitates robust detection performance even when one of the input modalities i.e. audio or video, is missing. 

Another work \cite{R144} utilizes monomodal datasets (visual-only or audio-only deepfakes) for training, rather than relying on multimodal deepfake data. This approach extracts audio-visual features over time and analyzes them using time-aware neural networks, capitalizing on the inconsistencies both within and across modalities to improve detection performance. The proposed method is evaluated on unseen multimodal deepfakes, allowing it to evaluate the robustness of the detector without requiring multimodal training data. 

To address the generalization issue in deepfake detection, a person-of-interest (POI) forgery detector is proposed in \cite{R142}, to learn each individual’s most discriminative identity features through a contrastive learning framework. This approach enables robust forgery detection without reliance on artifacts or manipulation traces generated using deepfake techniques.

Finally, \cite{R139} proposes a multimodal network with a web interface for real-time video forgery detection. This system integrates multiple data modalities to boost detection accuracy and provides a user-friendly, accessible platform for real-time analysis.

\section{Datasets for Audiovisual Deepfake Detection}
\label{sec:Datasets for Audiovisual Deepfake Detection}

To effectively detect audiovisual deepfakes, a dataset with multiple manipulation methods, real-life scenarios, and increasingly sophisticated deepfake generation techniques is required to provide a valuable resource for training, validation, and testing detection systems. This section discusses existing widely used datasets that were developed to aid research in audiovisual deepfake detection. The statistics of these datasets are shown in Table \ref{table:datasets}.

\begin{table*}[h!]
\centering
\caption{Comparison of existing audiovisual deepfake datasets.}
\small

\begin{tabular}{|c|c|c|c|r|r|r|}
\hline
\textbf{Dataset} & \textbf{Year} & \textbf{Manipulation} & \textbf{Subjects} & \textbf{Real} & \textbf{Fake} & \textbf{Total} \\ \hline\hline
DFDC \cite{R6} & 2019 & AV &  960 & 23,654 & 104,500 & 128,154 \\ \hline
FakeAVCeleb \cite{R93} & 2021 &  AV & 500 & 500 & 19,500 & 20,000 \\ \hline
LAV-DF \cite{R9} & 2022 & AV & 153 & 36,431 & 99,873 & 136,304 \\ \hline
AV-Deepfake1M \cite{R11} & 2023 & AV & 2,068 & 286,721 & 860,039 & 1,146,760 \\ \hline
AV-PolyGlotFake \cite{R145} & 2023 & AV & - & 766 & 14,472  & 15,238 \\ \hline
\end{tabular}
\label{table:datasets}
\vspace{-2mm}
\end{table*}

\subsection{DFDC (DeepFake Detection Challenge)}
A preview dataset containing 5,214 videos was released in October 2019, followed by the full dataset in December 2019, containing 119,154 videos, each 10 seconds long, with 486 unique subjects, additional information is available in Table \ref{table:datasets}. The DeepFake Detection Challenge (DFDC) \cite{R6} dataset is a large-scale publicly accessible dataset released by Facebook AI to promote research and development in deepfake detection methods. 

This large-scale dataset was recorded in natural settings using high-resolution cameras without professional lighting or makeup, and was collected from 3,426 paid subjects. It includes 1,000 deepfake videos generated using various techniques, including GAN-based and non-learning methods, for each genuine video. The dataset also includes various types of deepfakes, such as faceswap, face reenactments, and full-body deepfakes, and manipulations in audio, visual, or both audio and visual streams. This dataset stands out for the diversity and variability of deepfake generation methods, focusing on audiovisual components, high-quality videos, and extensive use in benchmarking.

\subsection{FakeAVCeleb}
The FakeAVCeleb dataset \cite{R93} is a comprehensive collection of audiovisual deepfake recordings of celebrities designed to develop, advance, and evaluate deepfake detection methods. Developed by researchers at Sungkyunkwan University, South Korea, the primary purpose is to assist the scientific community in benchmarking deepfake detection models, particularly those that examine both visual and auditory modalities.

The dataset consists of 500 original videos, each approximately 30 seconds long, featuring a variety of celebrities from the sports, politics, music, and film industries, including Barack Obama, Donald Trump, and Kim Kardashian. Based on 500 original videos, 19,500 manipulated samples were generated through various manipulation techniques (such as Faceswap \cite{R1} and Wav2lip \cite{R70}) and real-time voice cloning (RTVC) (such as SV2TTS \cite{R5}). The dataset consists of 500 subjects, of which 470 subjects' videos are used for training and the remaining 70 subjects' videos are used for testing. The FakeAVCeleb dataset is known for its multimodal manipulation, which makes it more challenging and relevant than synthetic datasets that only contain unimodal manipulation. This challenging dataset can make significant contributions to academic research and practical applications in the fields of cybersecurity and media forensics, and can be used to benchmark deepfake detection, multimodal forgery detection, and human perception research.

\subsection{LAV-DF (Localized Audio Visual DeepFake)}
The Localized Audio Visual DeepFake (LAV-DF) dataset \cite{R9} is a multimodal dataset released by a group of researchers in 2022 to advance research on deepfake detection, with a particular focus on locating alterations in audio and visual data. It contains 136,304 videos, including 36,431 real videos of 153 unique subjects and the corresponding 99,873 fake videos. LAV-DF features comprehensive audio and visual deepfakes with targeted content-driven manipulations (utilizing content-driven reenactment and text-to-speech methods) designed to more accurately replicate authentic speech and expressions. When evaluated on this dataset, the deepfake detection model not only needs to identify the deepfake content, but also accurately locate the exact fake region.

\subsection{AV-Deepfake1M}
Released in 2023, the large-scale AV-Deepfake1M dataset \cite{R11} is carefully designed to facilitate the development of robust deepfake detection and localization techniques. The dataset contains more than 1 million videos featuring 2000+ subjects, encompassing audio, visual, and audiovisual manipulations. It provides a challenging benchmark for evaluating the efficacy of current detection models and encourages innovation to effectively mitigate the threat of deepfakes to media authenticity. 

\subsection{PolyGlotFake}
PolyGlotFake is a novel comprehensive multilingual and multimodal dataset released in 2023, developed to advanced techniques in deepfake detection. PolyGlotFake spans seven languages including English, French, Spanish, Russian, Chinese, Arabic, and Japanese and incorporates a diverse set of manipulation techniques including advanced text-to-speech, voice cloning, and lip-sync technologies. The dataset comprises over 15,000 videos, encompassing both audio and visual manipulations, making it a valuable asset for driving advancements in holistic deepfake detection research.

\section{Performance Metrics and Evaluation}
\label{sec:Performance Metrics and Evaluation}
Evaluating the performance of deepfake detection systems relies heavily on robust evaluation metrics that provide measurable insights into the model's accuracy, robustness, and generalizability, further allowing comparison of various detection methods and progress observations. This section presents common metrics for evaluating audiovisual deepfake detection systems.

\subsection{Accuracy-Based Metrics}
\subsubsection{True Positive Rate (TPR), False Positive Rate (FPR), True Negative Rate (TNR), and False Negative Rate (FNR)}
Such metrics reflect a model's ability to correctly classify authentic content versus manipulated content. True Positive Rate (TPR), also known as recall or sensitivity, shows how effectively a model detects deepfakes, and is defined as
\begin{equation}
\text{TPR} = \frac{TP}{TP + FN},
\end{equation}
where $TP$ and $FN$ denote True Positive (fake instances correctly detected as fake) and False Negative (fake instances incorrectly detected as real), respectively.
False Positive Rate (FPR) measures the probability that an authentic instance will be incorrectly identified as a deepfake, and is defined as
\begin{equation}
\text{FPR} = \frac{FP}{FP + TN},
\end{equation}
where $FP$ and $TN$ denote False Positive (real instances incorrectly detected as fake) and True Negative (real instances correctly detected as real), respectively.
True Negative Rate (TNR) represents the rate of instances correctly detected as real with respect to all existing real instances, and is defined as 
\begin{equation}
\text{TNR} = \frac{TN}{FP + TN}.
\end{equation}
False Negative Rate (FNR) represents the rate of instances incorrectly detected as real with respect to all existing fake instances, and is defined as
\begin{equation}
\text{FNR} = \frac{FN}{FN + TP}.
\end{equation}
These metrics are often combined to provide a comprehensive assessment of detection performance.

\subsubsection{Precision, Recall, and F1-Score}
Precision measures how accurately a model identifies fake instances. It is the ratio of $TP$ to the sum of $TP$ and $FP$:
\begin{equation}
\text{Precision} = \frac{TP}{TP + FP}.
\end{equation}
Recall (also known as sensitivity or TPR) evaluates the capability of a model to effectively identify all fake instances:
\begin{equation}
\text{Recall} = \frac{TP}{TP + FN}
\end{equation}
F1-Score is the harmonic mean of precision and recall: 
\begin{equation}
\text{F1-Score} = 2 \times \frac{\text{Precision} \times \text{Recall}}{\text{Precision} + \text{Recall}}.
\end{equation}
This metric is particularly valuable in cases where the dataset is imbalanced (one category is more prominent than the other).

\subsubsection{Accuracy}
Accuracy is a widely used evaluation metric for classification tasks to analyze the overall performance of a model. It is the ratio of correct predictions ($TP$ and $TN$) to the total number of instances in the dataset:
\begin{equation}
\text{Accuracy} = \frac{TP + TN}{TP + TN + FP + FN}.
\end{equation}

\subsection{ROC and AUC}
\subsubsection{Receiver Operating Characteristic (ROC) Curve}
The Receiver Operating Characteristic (ROC) curve visually demonstrates how the model balances the true positive rate and false positive rate by displaying TPR versus FPR under different thresholds. The ROC curve in the upper left corner represents the model with the highest TPR and lowest FPR.  

\subsubsection{Area Under the Curve (AUC):}
Area Under the Curve (AUC) refers to the area under the ROC curve. An AUC closer to 1 represents better performance, while an AUC of 0.5 indicates a lack of discrimination, similar to random guessing.

\section{Human Perception and Psychological Impact in Audiovisual Deepfakes}
\label{sec:Human Perception and Psychological Impact in Audiovisual Deepfakes}
Human perception plays a crucial role in spotting and analyzing the authenticity of audiovisual deepfakes. The complexity of human sensory processing allows individuals to detect subtle discrepancies, especially in nuances that automated systems may miss, such as facial expressions, tone of voice, lip synchronization, and synchronized audio and visual aspects. Human subjective judgment and detection proficiency can enhance the performance of automated detection systems.

\subsection{Human Perception Studies in Deepfake Detection}
\subsubsection{Human Detection Capabilities and Challenges}
Despite advances in audiovisual deepfake generation \cite{R93}, humans inherently possess visual and auditory cognitive or perceptual skills that help detect subtle inconsistencies. However, research in \cite{R117, R118, R119, R122} shows that, contrary to expectations, a significant number of individuals are unexpectedly vulnerable to high-quality deepfakes, especially when subtle audiovisual alterations coincide with realistic behavioral signals \cite{R116}.  

Studies investigating humans' ability to detect deepfakes \cite{R116, R120, R121, R123, R147} have considerable variation with factors such as quality of manipulation, type of manipulation, familiarity with media figures, digital literacy, and age group \cite{R125}. Humans often rely on their own instincts, nonverbal cues such as unnatural eye blinks, discrepancies in lip synchronization or inconsistencies in audio-visual synchronization \cite{R124} that advanced detection models may not catch.

\subsubsection{Factors Influencing Human Detection Performance}
Existing knowledge, cognitive biases, experience and expertise, contextual understanding, and emotional intelligence are several important factors that affect human perception and judgment of the authenticity of audiovisual content. Understanding these biases, the impact of prior knowledge, and having professional experience and expertise are critical to making nuanced decision and are essential to improving the design of detection tools and their validity and reliability \cite{R123}. The uncanny valley effect, where near-perfect replicas appear perturbing and raise the uncertainty leading to deepfake detection, is particularly relevant for identifying deepfakes that almost, but not entirely, mimic real human behavior. However, humans often miss manipulation signs due to limited attention or fatigue, which is increased by prolonged exposure to digital content \cite{R121} \cite{R126}. 

\subsection{Comparing Human and Machine Detection}

\subsubsection{Comparison of the Capabilities of Human and AI Models}
AI algorithms for audiovisual deepfake detection surpass human perception in detecting audiovisual deepfakes \cite{R116} \cite{R126} \cite{R127}. Unlike humans, these detection algorithms can analyze large amounts of content quickly, precisely, and consistently. Many AI models are highly scalable and adaptable and free of biases that humans may have. However, humans possess unique abilities in contextual understanding and subjective interpretation \cite{R123}. Social experience enables people to interpret nuances and adapt to complex, unforeseen scenarios, which helps intuitively detect subtle irregularities in expressions or unnatural tones that may be lost without explicit feature engineering. This comparative strength highlights the promising benefits of integrating human and AI capabilities in practical applications \cite{R128}. 

\subsubsection{Challenges in Automating Human Intuition}
Translating human subjective intuition into automated objective measurable features remains inherently difficult. The human ability to recognize micro-expressions, contextual inconsistencies, or subtle unnatural behaviors in manipulated content does not always translate into measurable features in algorithms; this intuition is abstract. Some methods show potential to use physiological data to bridge this gap,  such as monitoring pupil dilation through eye tracking or measuring reaction times to audiovisual cues. Embedding insights from human intuition into machine learning frameworks may improve the accuracy of detection models \cite{R128}.


\section{Current Challenges and Future Directions}
\label{sec:Current Challenges and Future Directions}
Generalization remains a significant challenge for audiovisual deepfake detection. Existing audiovisual models perform well on seen datasets and deepfake techniques; however, their performance degrades significantly on unseen datasets and new manipulation techniques. Additionally, videos shared on online social media platforms often come with complexities including illumination variation, non-frontal faces, far frontal faces, occluded faces or lips, background clutter, multiple speakers, compression artifacts, multiple camera angles, and environmental noises. These factors pose challenges to multimodal forgery detection systems, which are typically trained using clean, preprocessed, and biased datasets.  

To address these generalization issues, it is crucial to develop diverse datasets that contain samples generated by a variety of state-of-the-art deepfake techniques and are free from gender and racial bias. The approach to dataset creation and development of multimodal deepfake detection systems must be iterative. As new deepfake generation techniques emerge, it will be important to include more generated video samples into existing datasets. This continuous updating requires regular model updates to effectively combat novel video forgery techniques. Furthermore, innovative approaches such as novel multimodel feature fusion and unsupervised or self-supervised learning-based foundation models can be leveraged to learn robust audiovisual features and potentially improve performance across datasets and manipulations. Techniques such as domain adaptation and transfer learning can further improve the generalization ability of multimodal deepfake detectors.  

The scalability of current audiovisual forgery detection systems is also challenging. Deploying these systems on social media platforms that host millions of videos and online users requires computational efficiency to ensure real-time performance. Audiovisual forgery detection models process two data streams simultaneously. This bimodal approach, combined with preprocessing and high data dimensionality in the forward pass, requires substantial resources to effectively analyze visual and acoustic features. To facilitate authenticity checking of visual and acoustic streams, lightweight video and audio feature extractors and resource-efficient neural networks are needed. For real-time detection systems, techniques such as pruning, quantization, and knowledge distillation must be employed to develop resource-efficient multimodal forensic methods. By optimizing these processes, we can enhance the scalability and effectiveness of audiovisual forensic systems in dynamic online platforms.

Since deepfake manipulation primarily targets facial and speech data, it raises serious privacy and ethical issues when collecting personal data to train data-driven end-to-end models. Consent must be ensured and ethical standards must be adhered to, while strong privacy protections must be implemented. These steps are critical to preventing potential harm while effectively addressing the challenges of forgery detection. Furthermore, it is crucial to incorporate model transparency and specific preprocessing steps before analyzing facial, lip, and speech biometric features. This approach not only enhances privacy protection but also ensures that individual rights are respected throughout the detection process. By prioritizing these ethical considerations, we can work toward developing more responsible and effective deepfake detection systems that protects individual privacy.

\section{Conclusions}
\label{sec:Conclusion}
The rise of audiovisual deepfakes has eroded people's trust in digital content, making seeing fall short of believing. The ubiquity of audio and visual manipulation tools, coupled with the rapid and effortless spread of fake content through social media platforms, greatly increases the risk of abuse. In line with this trend, multimodality has also revolutionized the field of deepfake detection, showcasing cutting-edge performance and unparalleled versatility in identifying forged content across different modalities. This survey paper highlights recent trends in multimodality in forgery detection and provides an in-depth look at the complexities and advances in multimodal or audiovisual deepfake detection technologies. We comprehensively explore existing studies that use multimodality, especially audio and visual streams via AI algorithms, to detect forgery in videos. We also provide information on large-scale and high-quality datasets suitable for multimodal deepfake detection, which are considered key factors in advancing the capabilities of detection methods. This survey also sheds light on the unique perspective of integrating human perceptual cues into algorithmic models, enabling these models to improve detection efficacy and strengthen defenses against audiovisual forgeries. We also delve into open issues and future directions to provide a roadmap for further research. This survey is expected to inspire researchers to incorporate multimodality into deepfake detection, providing them with important insights into how multimodality can contribute to multimedia forensic tasks.

\bibliographystyle{IEEEtran}
\bibliography{References.bib}

\end{document}